\documentclass[
]{ceurart}

\usepackage{listings}
\usepackage{graphicx}
\usepackage{subcaption}
\usepackage{array} 
\usepackage{booktabs} 
\usepackage{longtable} 
\usepackage{multirow} 
\usepackage{pdflscape} 
\usepackage{ragged2e}
\usepackage{adjustbox}
\usepackage{tabularx}
\usepackage{rotating}
\usepackage{siunitx}
\usepackage{longtable}
\usepackage{xurl}
\usepackage[capitalise]{cleveref} 

\usepackage{sudocq}
\newcommand{\ontterm}[1]{\texttt{#1}}
\newcolumntype{C}[1]{>{\Centering\arraybackslash}m{#1}}
\newcolumntype{L}[1]{>{\RaggedRight\arraybackslash}m{#1}}
\newcolumntype{Y}{>{\RaggedRight\arraybackslash}X}
\newcolumntype{P}[1]{>{\RaggedRight\arraybackslash}p{#1}}

\begin{document}

\copyrightyear{2026}
\copyrightclause{Copyright for this paper by its authors. Use permitted under Creative Commons License Attribution 4.0 International (CC BY 4.0).}
\conference{6th International Workshop on Scientific Knowledge: Representation, Discovery, and Assessment, Oct 2026, Bari, Italy}


\title{SUDO: From Scientific Publications to Global Scholarly Discourse}


\author[1]{Nipun D. Pathirage}[%
orcid=0000-0002-9313-0925,
email=pathin@rpi.edu,
]
\address[1]{Rensselaer Polytechnic Institute, Troy, NY, USA}

\author[1]{Henrique Santos}[%
orcid=0000-0002-2110-6416,
email=oliveh@rpi.edu,
]

\author[1]{Deborah L. McGuinness}[%
orcid=0000-0001-7037-4567,
email=dlm@rpi.edu,
]


\begin{abstract}
Scholarly Knowledge Graphs (SKGs) support literature search, scientific discovery, and agent-assisted research. Yet, most approaches to SKG construction pre-specify the knowledge unit to be represented, typically named scholarly entities or contribution-level statements, and retain only the publication content expressible through that unit. This choice produces two dominant yet distinct paradigms: concept-centric SKGs that preserve entities and local relations, and statement-centric SKGs that capture high-level discourse in sentences. In both cases, the graph reflects a chosen extraction target rather than how a paper constructs and communicates knowledge. We introduce SUDO, the Scholarly Upper Discourse Ontology, which defines how knowledge elements are organized and related to form a paper’s research narrative, namely its scholarly discourse. Using linguistic principles that generalize across fields and tasks, SUDO enables each publication’s scholarly discourse to be constructed from the bottom up as a coherent, source-grounded representation. Applied consistently across papers, the same structure connects their research narratives into a global scholarly discourse without discarding local context or provenance. We demonstrate SUDO's capabilities across a comprehensive suite of scholarly knowledge-modeling use cases and performance improvements on a downstream Large Language Model (LLM)-based scientific QA task.
\end{abstract}


\maketitle

\section{Introduction}

Scholarly knowledge representation seeks to make scientific publications machine-actionable. Scholarly Knowledge Graphs (SKGs) operationalize this goal, supporting literature search~\cite{p_semschol,p_openalex}, knowledge discovery~\cite{p_sci_disc_review}, scientific question answering~\cite{d_sciqag}, and, increasingly, LLM- and agent-assisted scientific workflows~\cite{p_kggen,p_nlp_akg}. Existing SKGs largely follow two dominant paradigms: concept-centric~\cite{p_ai_kg,p_cs_kg} and statement-centric modeling~\cite{p_semsur,p_nlp_akg,p_orkg}. Both structures select aspects of publication knowledge, enabling scalable indexing, retrieval, comparison, and aggregation across large scholarly corpora.

Concept-centric SKGs represent scholarly knowledge through named entities and research artifacts, capturing their semantic categories and relations while leaving their roles within a publication's local context underspecified~\cite{p_semschol,p_ai_kg,p_cs_kg,p_cs_kg2}. For example, when the word ``\texttt{attention}" appears in a paper, a concept-centric SKG such as CS-KG~\cite{p_cs_kg2} may represent it as an instance of the \texttt{Method} category, but without capturing its role in that paper, whether as part of the proposed approach, a theoretical motivation, or a baseline. As a result, \texttt{attention} is represented uniformly across papers, obscuring the differences in how it is applied, adapted, or interpreted in each context. These are differences that are crucial for understanding how the concept develops and shifts across the literature, a central aspect of scholarly discourse.

Statement-centric SKGs provide a complementary view by representing higher-level scholarly assertions (e.g., research problems, approaches, evidence, and claims) and relating them to reconstruct a publication’s research argument~\cite{p_semsur,p_orkg,p_disc_skg}. Yet this abstraction creates three limitations. First, statements provide weak anchors for integration across publications, which typically depends on recurring concepts and shared scholarly nomenclature~\cite{p_semschol}. Second, because these ontologies predefine the scholarly aspects and statement types they capture, they represent only selected parts of a paper, omitting local details from methodology and evaluation as well as explicit cross-paper discourse in related work~\cite{p_nlp_akg}. More fundamentally, their statement types are designed to model scholarly functions, not to correspond to directly expressed units in the source. Instantiating them may therefore require combining information distributed across several passages or deriving a higher-level abstraction not stated in any one location~\cite{p_nlp_akg}. This weakens source grounding and is particularly problematic for LLM-based knowledge graph (KG) construction, where plausible statement-level descriptions can be generated without clearly identifiable textual support.

These limitations reveal a broader representational gap in which existing SKGs specify in advance which knowledge should be extracted, rather than how scholarly knowledge is built within a publication. Within a paper, knowledge is organized from the bottom up: established scholarly nomenclature anchors the work in prior knowledge; clauses and sentences use and relate that nomenclature, introduce new terms, and express pragmatics; and their progression across sentences and sections develops the paper’s research contribution. This organization constitutes its local scholarly discourse. Modeling that discourse preserves both the structural context and source grounding of knowledge before it is transformed into domain- or task-specific abstractions.

We operationalize this bottom-up view in SUDO, the Scholarly Upper Discourse Ontology. It draws on the theoretical linguistics of Jackendoff’s conceptual semantics, developed in the 1980s, which models how linguistic expressions map to structured conceptual representations of meaning~\cite{p_concept_sem}. SUDO extends this insight from individual expressions to scholarly publications, taking each publication’s lexical, sentential, and discourse organization as the basis of its knowledge representation: concepts are introduced and related within local linguistic contexts; discourse relations organize these expressions into the publication’s research argument; and recurring concepts connect publication-level arguments across the literature to form global scholarly discourse. Because these structures reflect general properties of language rather than field- or task-specific extraction categories, SUDO provides a representation that can be applied across domains and tasks.

We evaluate SUDO by constructing an SKG covering the International Semantic Web Conference's (ISWC) main-track publications from 2012 to 2025, assessing its support for a comprehensive collection of scholarly knowledge-modeling use cases, and demonstrating performance gains on downstream LLM-based scientific question answering. This work contributes a new computational-linguistic view of scholarly knowledge representation, its realization in the SUDO ontology, and the resulting ISWC scholarly knowledge graph.

\section{Related Work}
\subsection{The Data-Source Paradigm and Information Loss}

Formal scholarly knowledge representation has developed over nearly three decades, with increasing emphasis on scalable construction and access. Early systems such as ScholOnto~\cite{p_scholonto} represented scholarly claims through expressive, manually constructed ontologies. Although these representations captured rich scholarly semantics, their reliance on author or expert input limited their adoption at scale. Subsequent work therefore shifted toward automated and centralized scholarly infrastructures, including ArnetMiner~\cite{p_arnetminer}, AMiner~\cite{p_aminer}, the Microsoft Academic Graph (MAG)~\cite{p_mag}, and OpenAlex~\cite{p_openalex}. By integrating metadata, citation links, author identities, and topical indicators across large publication collections, these web-scale SKGs provided substantial value for literature discovery, filtering, bibliometric analysis, and scholarly search.

Advances in named-entity recognition and relation extraction enabled concept-centric SKGs, such as Semantic Scholar~\cite{p_semschol}, AI-KG~\cite{p_ai_kg}, and CS-KG~\cite{p_cs_kg,p_cs_kg2}, to be constructed automatically over web-scale publication collections and integrated into infrastructures such as MAG~\cite{p_mag2} and AMiner~\cite{p_aminer}. By linking recurring concepts and relations across publications, these graphs naturally support cross-publication knowledge exploration. However, they provide limited information about the role a concept plays within a specific paper. For example, whether it denotes a research problem, method, experimental variable, or prior work. This limits search and analysis that depend on publication-specific meaning and argumentative context, and makes concept-centric representations less compatible with LLM-based workflows that require connected, context-rich information.

Statement-centric SKGs were introduced to represent this publication-level information more explicitly. Systems such as ORKG~\cite{p_orkg}, SemSur~\cite{p_semsur}, and NLP-AKG~\cite{p_nlp_akg} model research contributions, experimental descriptions, or other scholarly information as natural-language statements. Compared with concept-centric graphs, these representations retain more of the meaning conveyed in the source publication and are therefore better suited to describing what a study investigates, proposes, observes, or concludes. Their construction, however, is more complex, particularly when statements must be extracted or generated automatically from information distributed across a paper. Consequently, statement-centric SKGs have not yet been widely adopted across web-scale publication collections.

Existing statement-centric SKGs typically represent only selected dimensions of scholarly content, such as research arguments~\cite{p_deligent}, contributions~\cite{p_orkg,p_nlp_akg}, or experiments~\cite{p_semsur}. They therefore provide only partial coverage of a publication, commonly model statements as isolated units without preserving the discourse relations that connect information across the paper, and, particularly in LLM-based construction, must ensure that generated statements remain faithful to and traceable to their source passages. These limitations have constrained their adoption in scientific question answering~\cite{d_spiqa,d_qasper,d_sciqag}, knowledge discovery~\cite{p_ai_scien,p_ai_scien2,p_alpha_evolve,p_alien_space}, and literature-review workflows~\cite{p_litsearch}, which continue to rely largely on full-text representations even when operating over web-scale publication collections. To bridge this gap, we introduce SUDO as a source-grounded representation for structurally modeling complete publications, connecting their statements through scholarly discourse, and supporting LLM-based workflows over web-scale scholarly knowledge graphs.

\subsection{Speech Acts and the Positioning of SUDO}

Linguistic theories have long examined how human expressions encode meaning and relate to one another within larger communicative structures~\cite{p_illocutionary,b_dst}. Rhetorical Structure Theory (RST), for example, models relations between discourse units and distinguishes their functional roles, such as central statements (\textit{nuclei}) and supporting statements (\textit{satellites})~\cite{p_rst}. Broader discourse theories extend these local relations to explain how individual statements collectively form coherent discourse~\cite{b_dst,p_dst}, while theories of illocution and commitment characterize the communicative intent an utterance expresses~\cite {p_illoc_logic,p_illocutionary,p_commitment_act}. Building on these foundations, Speech Act Theory provides a well-established framework for analyzing how utterance-level intentions contribute to conversation progression~\cite {p_speach_act}.

We bring these established linguistic foundations into scholarly knowledge modeling to address the limitations discussed above. SUDO is designed to structurally represent an entire publication while preserving its broader discourse organization. From this perspective, the global body of scientific literature can be understood as a longitudinal, multi-party dialogue. Within this dialogue, an individual paper functions as a complex linguistic artifact (a formal scholarly utterance) that responds to, builds upon, or challenges prior work through citations and shared conceptual vocabularies.

\section{Conceptual semantic view of scholarly knowledge} \label{sec:linguistic_view}

\begin{figure}[!t]
    \centering
    \includegraphics[width=0.98\linewidth]{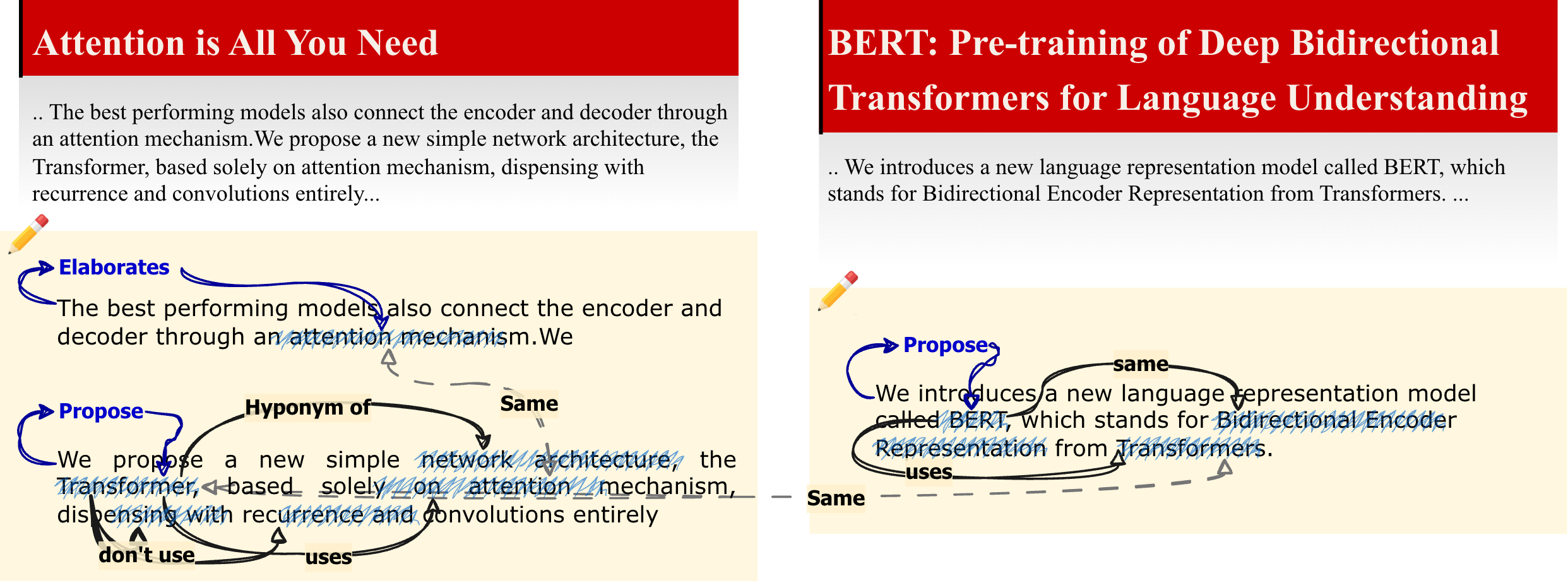}
    \caption{Illustration of how humans understand a publication's meaning and its relationships to other scholarly works from its linguistic structure.}
    \label{fig:running_example}
\end{figure}

We use the example in~\cref{fig:running_example} to illustrate the linguistic view adopted in this work. Consider the following sentence from the abstract of \texttt{Attention Is All You Need}~\cite{p_attention}: \textit{``We propose a new simple network architecture, the Transformer, based solely on attention mechanisms, dispensing with recurrence and convolution entirely.''} A reader recognizes that this sentence proposes \texttt{Transformer} as a new method, identifies it as a type of \texttt{network architecture}, and states that it is based on \texttt{attention mechanisms} rather than \texttt{recurrence} or \texttt{convolution}. The preceding sentence, \textit{``The best performing models also connect the encoder and decoder through an attention mechanism.''} plays a different role. It provides background for \texttt{attention mechanism} applications. Human readers infer these distinctions from linguistic cues expressed through noun phrases, clauses, lexical choices, and relations between sentences. Computational-linguistic theories such as Rhetorical Structure Theory~\cite{p_rst} and discourse-structure models~\cite{b_dst} seek to make such relations explicit.

This linguistic view provides a basis for modeling both local and global scholarly discourse. Consider a sentence from the abstract of the \texttt{BERT} paper~\cite{p_bert}: \textit{``We introduce a new language representation model called BERT, which stands for Bidirectional Encoder Representations from Transformers.''} Like the previous example, this sentence propose \texttt{BERT} as a new method. However, \texttt{Transformer} plays a different role here: it describes BERT's architectural basis rather than the principal contribution being introduced. Thus, the role of a concept within a publication is determined by how it participates in a clause and how that clause contributes to the broader discourse of the paper. Relations between concepts and clauses, and between clauses and publication-level functions, constitute the local discourse of a publication. Recurring concepts such as \texttt{Transformer} then connect these local discourse structures across papers, forming a global scholarly discourse.

This linguistic view allows the modeling of an entire publication by following how information is presented throughout the text. For example, when \texttt{Transformer} is introduced in one sentence and described further in later methodological sentences, the recurring named entity connects these distant clauses: the first proposes the approach, while the later clauses elaborate it with additional detail. The scattered information is thereby assembled into a coherent account of the approach without requiring a predefined contribution template. Importantly, lexical form varies far more than the linguistic properties that signal a scholarly function. An issue statement may be worded in many ways, but the cues marking it as the presentation of a research problem remain comparatively uniform. The same regularity holds for ideas, approaches, and claims, allowing any number of them to be represented consistently without prescribing how they must be expressed. Because each modeled element remains linked to the source text that expresses it, the resulting representation provides a fully source-grounded model of scholarly discourse.

Jackendoff’s \textbf{conceptual semantics} provides the theoretical mechanism for making this linguistic view computational~\cite{p_concept_sem}. It treats meaning as a structured conceptual representation assembled from a finite inventory of primitives and rules of combination. Lexical and syntactic forms remain distinct from this representation but are connected to it through explicit interface rules, allowing varied surface expressions to map to comparable linguistic structures. Within linguistic structure, \textbf{propositional structure} encodes predicate--argument content (roughly, who did what to whom) whereas \textbf{information structure} records how that content is situated communicatively, including distinctions such as topic and focus or given and new information. We adapt this architecture in SUDO: lexical and clausal analysis recovers scholarly objects and propositions, while discourse analysis assigns their scholarly functions and relations. These mappings operationalize the preceding linguistic view as a compositional knowledge representation constructed directly from publication text.

\section{SUDO}

\begin{figure}[!th]
    \centering
    \includegraphics[width=0.98\linewidth]{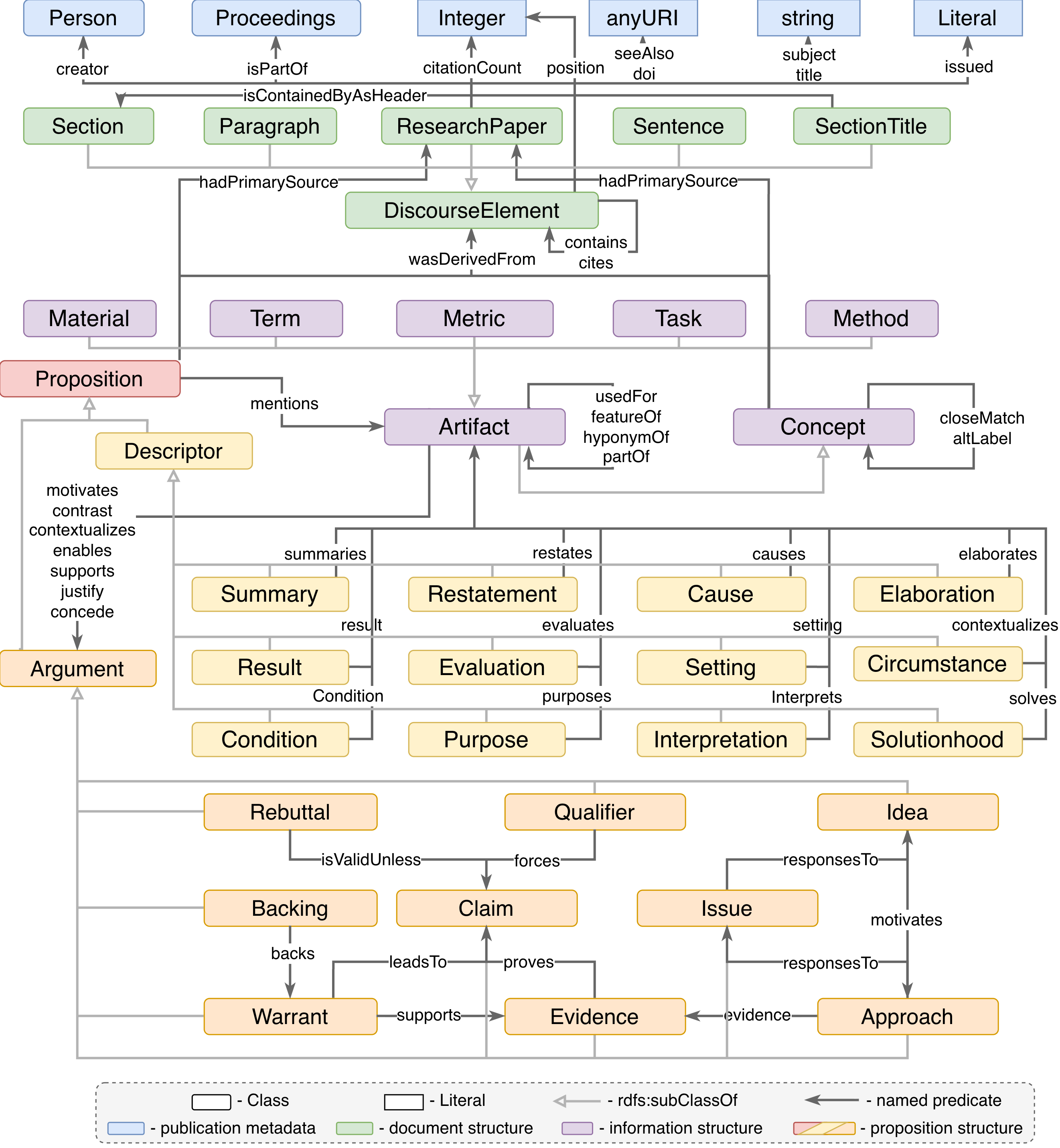}
    \caption{SUDO ontology diagram.~\ref{app:ontology_vocabulary}}
    \label{fig:sudo_ontology}
\end{figure}

\subsection{Overview}

The \textbf{Scholarly Upper Discourse Ontology} (SUDO) formalizes the conceptual semantic view of scholarly knowledge introduced in \Cref{sec:linguistic_view}. As illustrated in \Cref{fig:sudo_ontology}, SUDO represents a publication through its complementary \textit{information structure} and \textit{propositional structure}, together with the metadata and document structure required to locate this knowledge within the source publication. The propositional structure is further divided into \textbf{Arguments} and \textbf{Descriptors}. Arguments capture propositions that constitute the paper's research argument, whereas Descriptors capture propositions that describe, elaborate, or contextualize the scholarly concepts involved. This distinction draws on Jackendoff (2002)~\cite{p_found_lang} work on propositional segregation and Simone Teufel's argumentative-zoning theories~\cite{p_arg_zoning,p_arg_zone_thesis,p_toward_arg_zone}, which differentiate statements according to their roles in scholarly communication.

The following subsections describe how SUDO models and materializes the information and propositional structures in detail. We then show how these complementary structures are integrated to form the complete conceptual semantic representation of a scholarly publication.

\subsection{Information structure}

In SUDO, the information structure models the scholarly nomenclature through which a publication establishes what it discusses and organizes its propositions. It therefore represents noun-phrase expressions as named scholarly entities and captures the semantic relations among them. To remain applicable across fields, these entities are classified using broad scholarly semantic categories rather than field-specific subject matter. For this purpose, SUDO adapts the cross-domain SciERC scientific named entity recognition (NER) schema, using \textbf{Task}, \textbf{Method}, \textbf{Metric}, \textbf{Material}, and \textbf{Term}. Relations are defined at the same level of generality: \textbf{used-for}, \textbf{feature-of}, \textbf{hyponym-of}, \textbf{part-of}, \textbf{compare}, and \textbf{conjunction} capture functional, compositional, and taxonomic associations~[\cref{fig:sudo_ontology}]. Crucially, SUDO distinguishes a nomenclature’s textual realization from its canonical identity. Each noun-phrase occurrence is represented as an \textbf{Artifact}, preserving its source location, lexical form, abbreviation or coreferential expression, and context-specific semantic category. The corresponding \textbf{Concept} represents the canonical identity shared across occurrences. Artifact instances can subsequently be reconciled with Concepts through mappings such as \textbf{closeMatch}, while lexical alternatives are retained through \textbf{altLabel}. This separation preserves publication-specific usage while supporting concept alignment across the literature. Further definitions of the entity and relation types are provided in \Cref{app:concept_structure}.

\Cref{fig:sudo_example} materializes this design through the running \texttt{Transformer} and \texttt{BERT} example in \Cref{fig:running_example}. The occurrences \texttt{The Transformer} in \texttt{Attention Is All You Need} and \texttt{Transformers} in the \texttt{BERT} paper are represented as separate \texttt{Method} Artifacts linked to their respective source sentences. These lexical realizations remain distinct in their publication contexts, while \texttt{closeMatch} maps both to the canonical \texttt{Transformer} Concept. Their semantic relations contribute, through this canonical alignment, to a shared cross-publication concept structure. Because the Artifacts remain distinct, each can also participate in the \texttt{Approach} proposition that captures its role within the corresponding publication. The information structure thereby connects recurring nomenclature across papers without erasing the lexical and contextual distinctions required by SUDO’s propositional structure.

\subsection{Propositional Structure}

The \textbf{propositional structure} models how a publication turns the scholarly objects in its information structure into a research narrative. Each \textbf{Proposition} is anchored to a source clause and linked to the \textbf{Artifacts} that participate in it, preserving what the publication states about each locally realized concept. Some propositions perform the principal scholarly acts of the paper. For example, posing an issue, proposing an approach, presenting evidence, or making a claim, establishing argumentative backbones. Other propositions supply the semantic detail through which those acts are contextualized, qualified, or summarized, connecting locally expressed detail. This separation aligns with argumentative-zoning theories, which distinguish scholarly text by its rhetorical function within the publication~\cite{p_arg_zoning,p_toward_arg_zone}. Together, they represent both the publication-level research argument and the source-grounded information from which it is constructed.

\paragraph{Argument Structure.}  

The \textbf{Argument} structure models the propositions and relations that form a publication’s argumentative backbone and develop its research contribution. Indirect argumentation is a natural fit because this backbone follows the scientific method: an \textbf{Issue} motivates an \textbf{Idea} or \textbf{Approach}, its evaluation produces \textbf{Evidence}, and \textbf{Warrants} connect that Evidence to resulting \textbf{Claims}. The justificatory roles of Evidence, Warrant, and Claim follow Toulmin’s model of practical argumentation~\cite{p_toulmin}, as formalized in the Argument Model Ontology (AMO)~\cite{o_amo}. SUDO complements them with the Issue, Idea, and Approach roles used by SWiM/SIOC to adapt argumentation modeling to scientific discourse~\cite{p_deligent}. Rhetorical Structure Theory and discourse-structure models provide the linguistic basis for recognizing how these functions and their connections are expressed in text~\cite{p_rst,b_dst}. Together, these elements relate propositions through \textbf{responds-to}, \textbf{motivates}, \textbf{supports}, and \textbf{leads-to}~[\Cref{fig:sudo_ontology}], allowing each contribution to form its own argument structure within the publication.

\begin{figure}[!t]
    \centering
    \includegraphics[width=0.98\linewidth]{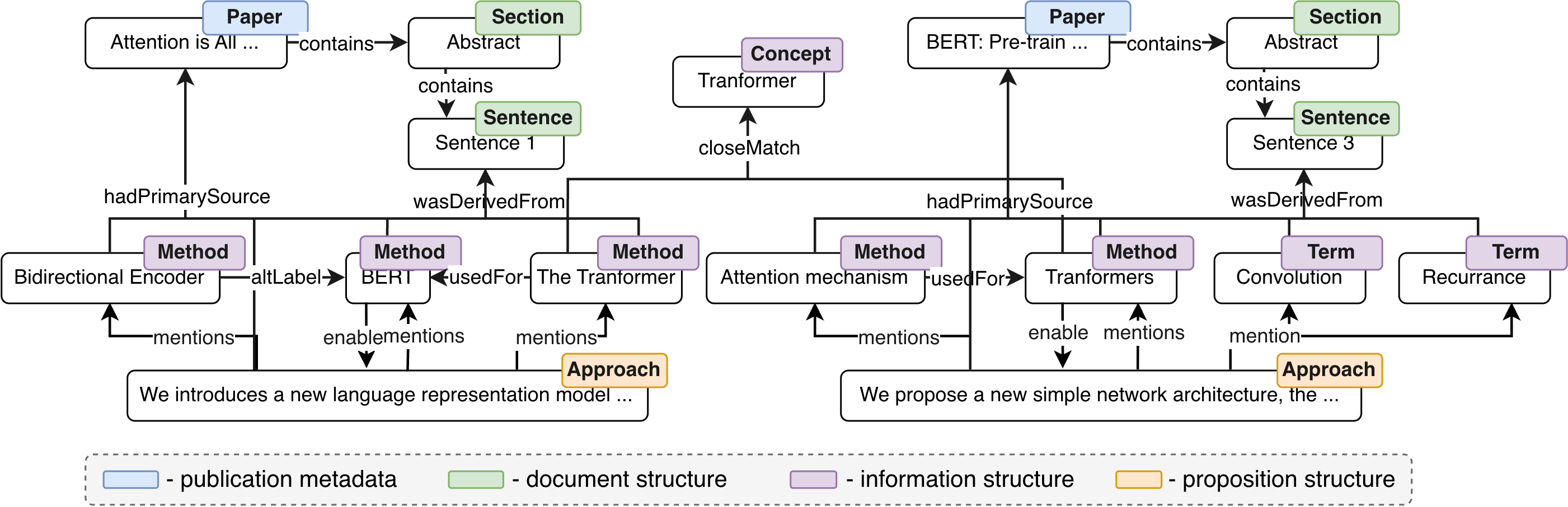}
    \caption{SUDO realization of the scenario introduced in \cref{fig:running_example}, emphasizing the constructed information and propositional structures while illustrating how their elements remain grounded in the source publications through document structure and metadata.}
    \label{fig:sudo_example}
\end{figure}

In \Cref{fig:sudo_example}, the proposal in \texttt{Attention Is All You Need} forms an \textbf{Approach} proposition centered on \texttt{Transformer}. The introductory statement in the \texttt{BERT} paper forms a separate Approach centered on \texttt{BERT}, with \texttt{Transformer} appearing as its architectural basis. Although the two Transformer Artifacts align with the same canonical Concept, their proposition-level connections preserve these different argumentative roles. Connecting each Approach to its associated Issues, Evidence, Warrants, and Claims then reconstructs the distinct research argument each publication develops.

\paragraph{Descriptor Structure.}  
The \textbf{Descriptor} structure models propositions that supply semantic detail about an Artifact or another Proposition. Rhetorical Structure Theory provides a natural framework by classifying interactions between rhetorical elements according to how they contribute to the overall statement~\cite{p_rst}. In \textit{subject-matter relations}, a rhetorical statement supplies semantic detail about another element; in \textit{presentational relations}, a rhetorical statement uses another element to advance its communicative purpose. Descriptors therefore correspond to RST’s subject-matter relations. RST defines fourteen such relations, of which \textit{sequence}, \textit{contrast}, and \textit{joint} are multinuclear and may involve more than two discourse units. Representing these relations through binary RDF/OWL object properties would require an additional n-ary reification pattern~\cite{p_owl_onto}. SUDO consequently adopts the remaining eleven binary relations, realizing each as a Descriptor type paired with a property that links the Descriptor to the Artifact or Proposition it describes~[\Cref{fig:sudo_ontology}].

Returning to the running example, later methodological clauses describing the architecture or internal layers of \texttt{Transformer} become Elaboration Descriptors. Through their shared Transformer Artifact, these Descriptors attach the dispersed methodological detail to the Approach that introduced the method, without treating each description as a separate contribution. Because every Descriptor corresponds to the clause that expresses it, the resulting detail remains source-grounded while contributing to the publication’s broader discourse.

\subsection{Conceptual Semantic View of Global Scholarly Discourse}

SUDO constructs a publication’s local scholarly discourse by joining the structures defined above. Artifact instances and their semantic relations form its publication-specific information structure. \textbf{Argument} propositions organize \textbf{Issue}, \textbf{Idea}, \textbf{Approach}, \textbf{Evidence}, \textbf{Warrant}, and \textbf{Claim} acts into one or more argumentative backbones, while \textbf{Descriptor} propositions attach semantic detail to those arguments and their Artifacts. Artifact--Proposition links bind the two structures by recording which locally realized nomenclature participates in each proposition. Global scholarly discourse is then constructed through canonical \textbf{Concepts}: \texttt{closeMatch} aligns recurring Artifacts across publications, connecting their surrounding proposition structures without merging the source-specific Artifacts themselves. 

Concept graphs can encode only relations anticipated by their predicate vocabulary. When a publication expresses a more nuanced connection, the graph must either omit it or reduce it to a generic predicate, making semantically distinct relations indistinguishable and losing how the concepts are actually connected in the source. Alongside discrete Concept--Concept relations, SUDO links Artifacts to the source Propositions in which they participate. These Propositions preserve the specific meaning underlying generalized Concept relations and capture connections for which no discrete relation is instantiated. Because this meaning remains accessible through the Proposition’s linguistic representation in latent semantic space, the Proposition acts as a semantic reification of the relation, a behavior we call \textbf{latent semantic reification}.

\subsection{Implementation}

SUDO was developed following the Ontology Development 101 methodology~\cite{b_onto_101} and has undergone three major revisions across 19 update cycles. Version~3.5.2 is the current stable release, with backward compatibility maintained throughout the 3.x series. SUDO extensively reuses established ontologies for publication metadata, document structure, and provenance, as detailed in \Cref{app:ontology_vocabulary}. Publication metadata is represented in its most general reusable form, allowing derivative ontologies to introduce more specialized models without losing information during mapping. The ontology is version-controlled through Git and published under the persistent W3ID namespace \texttt{https://w3id.org/twc/sudo/} in RDF/XML, OWL/XML, Turtle, N-Triples, and JSON-LD. Each release includes WIDOCO-generated documentation~\cite{p_widoco}, maintained competency questions, and an OOPS! evaluation for common ontology-modeling pitfalls~\cite{p_oops}.

\subsection{User Case Scenario}

To demonstrate SUDO’s utility for scholarly knowledge modeling, we construct a knowledge graph from 898 ISWC main-track papers published between 2012 and 2025, including their authorship, publication year, and track metadata. For each publication, triples describing publication metadata, document structure, local scholarly discourse, and provenance are stored in four dedicated named graphs, with provenance statements linking them into a unified publication representation. The complete ISWC corpus and resulting knowledge graph are publicly available through the resources listed in \Cref{app:online-resources}. Graph construction uses a neuro-symbolic pipeline whose methodology is treated separately, allowing this paper to focus on SUDO’s ontology design and evaluation.\footnote{The construction pipeline is described in a separate manuscript currently under review; its reference will be added once that work is published.} The competency-question evaluation in \Cref{app:cq} is conducted over this graph, with the complete question set provided in the appendix.

Using the same pipeline, we additionally construct and publish a SUDO knowledge graph for 413 publications from the QASPER benchmark~\cite{d_qasper}. This graph supports the LLM-based usability evaluation described in \Cref{sec:rq2}.

\section{Evaluation}

Because SUDO is intended both to structurally represent scholarly knowledge and to support its practical use, its evaluation must extend beyond any single dataset or task. We therefore combine two complementary forms of assessment: a use-case-based analysis of representational coverage and a downstream evaluation of practical utility in LLM-assisted scientific question answering. These assessments address the following research questions:

\begin{itemize}
\item \textbf{RQ1:} To what extent does SUDO structurally represent scholarly knowledge throughout a publication compared with existing SKG models?
\item \textbf{RQ2:} Does SUDO’s discourse-aware representation improve information access in LLM-based scientific question-answering workflows?
\end{itemize}

For each research question, we first define the corresponding evaluation objective and procedure, then report results and discuss the evidence-supported conclusions.

\subsection{RQ1: Structural Representation Coverage} \label{sec:rq1}

\begin{figure}[t]
    \centering

    \begin{subfigure}[t]{0.24\linewidth}
        \centering
        \includegraphics[width=\linewidth]{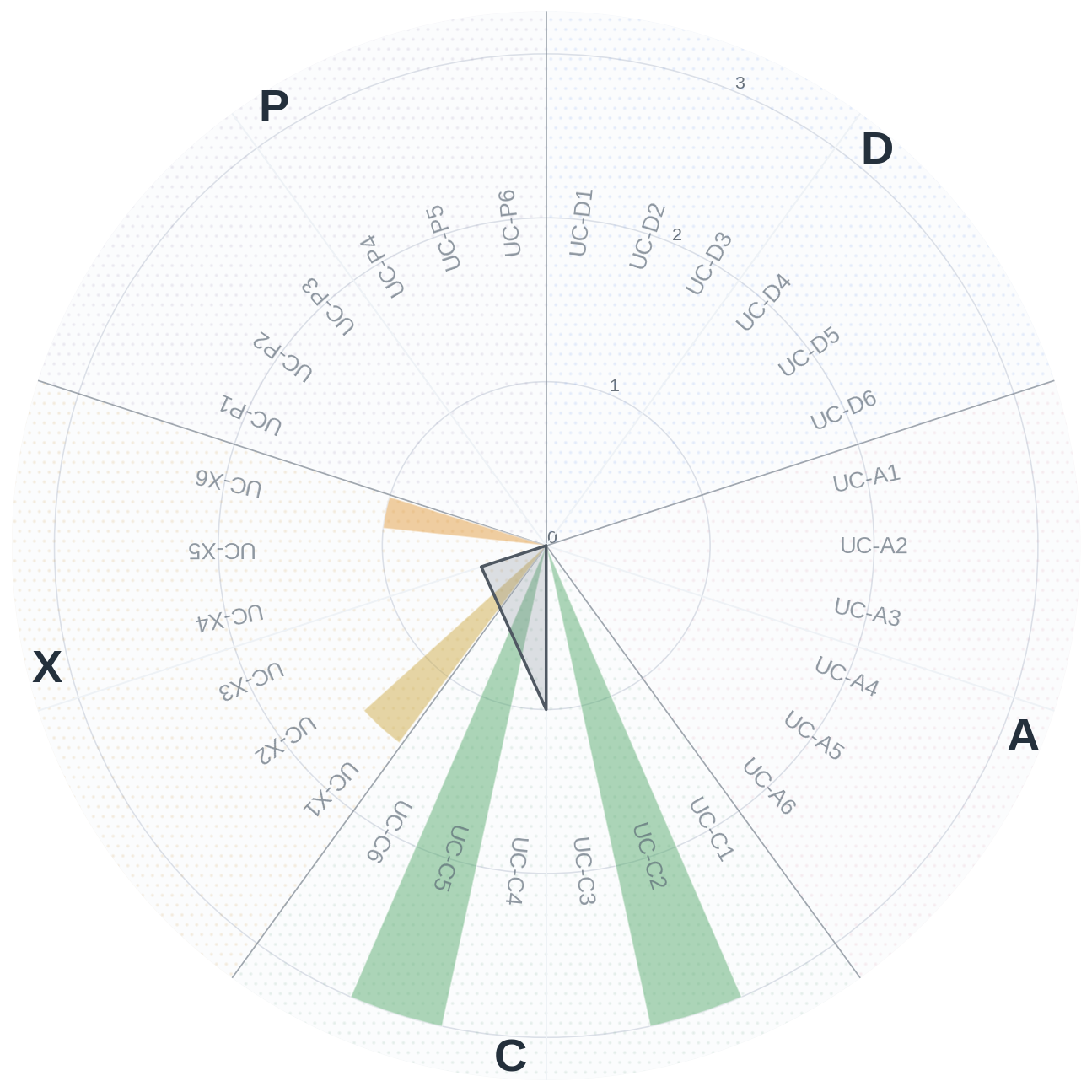}
        \caption{cs-kg}
        \label{fig:cskg_uc}
    \end{subfigure}
    \hfill
    \begin{subfigure}[t]{0.24\linewidth}
        \centering
        \includegraphics[width=\linewidth]{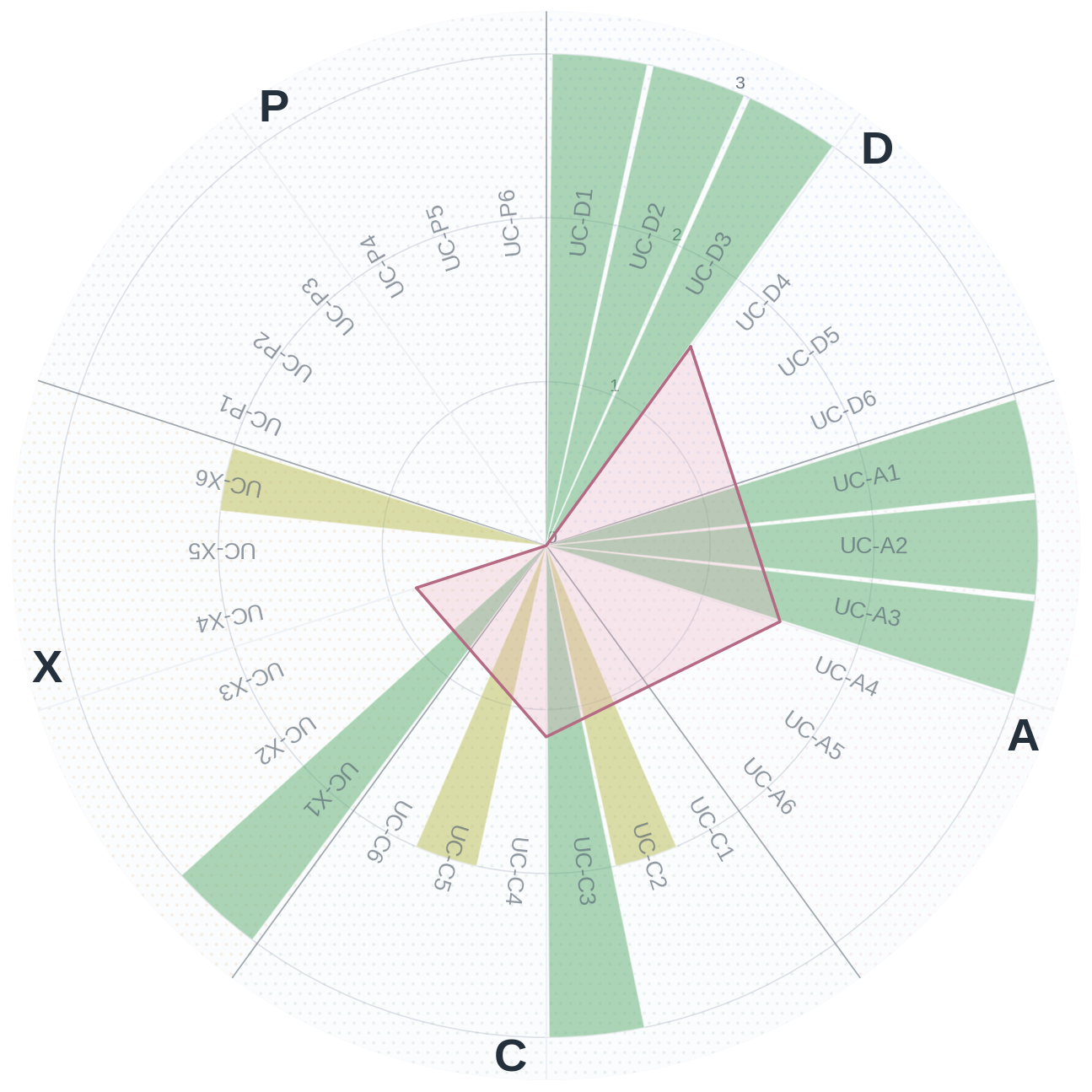}
        \caption{nlp-akg}
        \label{fig:nlp-akg_uc}
    \end{subfigure}
    \hfill 
    \begin{subfigure}[t]{0.24\linewidth}
        \centering
        \includegraphics[width=\linewidth]{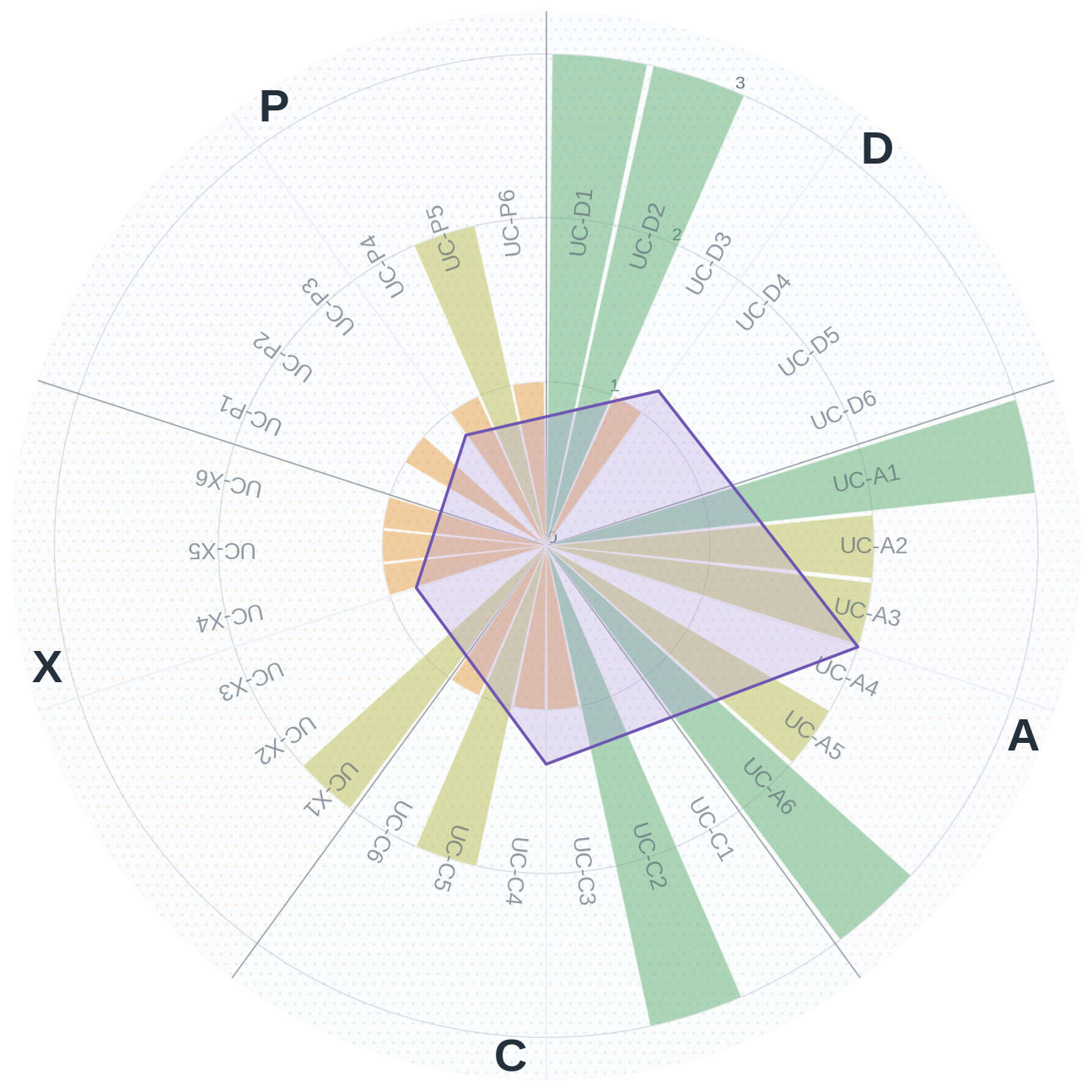}
        \caption{orkg}
        \label{fig:orkg_uc}
    \end{subfigure}
    \hfill 
    \begin{subfigure}[t]{0.24\linewidth}
        \centering
        \includegraphics[width=\linewidth]{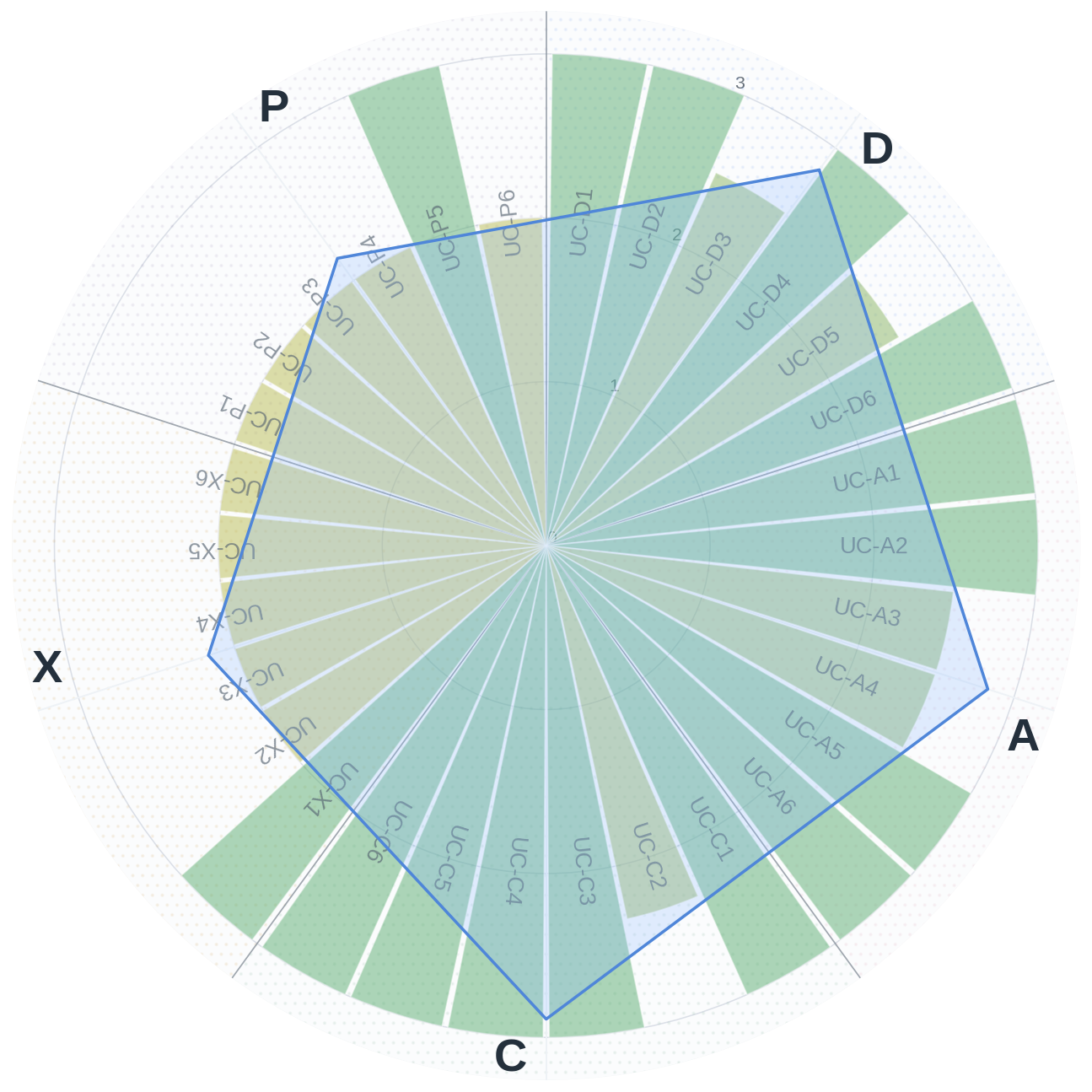}
        \caption{\textbf{sudo}}
        \label{fig:sudo_uc}
    \end{subfigure}
    
    \caption{
    Comparative use-case coverage across SKGs. Nightingale rose sectors show individual use-case satisfiability($S_{UC}$) on a four-level scale (0--3), with no bar for Level~0 and red, yellow, and green denoting Levels~1, 2, and 3, respectively; the overlaid radar plot shows family-level coverage computed from the mean $S_{UC}$. Sector labels denote the family IDs in \Cref{tab:publication-use-cases}, with use cases ordered clockwise from 1 to 6 within each family.
    }
    \label{fig:uc_analysis}
\end{figure}

Competency questions are the established means of testing whether an ontology or populated knowledge graph can satisfy its intended information needs. Because they are formulated for a particular schema and dataset, however, competency-question collections do not readily generalize across SKGs or support direct coverage comparisons. RQ1 therefore requires a shared evaluation framework rather than a SUDO-specific question set. We introduce a collection of ontology-independent scholarly knowledge-modeling use cases, each of which can be operationalized as a schema-specific competency question for every SKG under evaluation. This preserves a common information requirement while allowing each model to answer it through its own representational structure. We derived the collection by examining nearly two decades of scholarly knowledge-modeling research and published use cases, identifying five recurring families: \textit{publication metadata and document structure}, \textit{paper argument and discourse}, \textit{concepts and paper-local use}, \textit{cross-publication concept dynamics}, and \textit{paper-local pragmatic descriptions}. Each family contains six use cases, yielding 30 in total. The complete collection is provided in \Cref{tab:publication-use-cases}.

\begin{table}[t]
\centering
\caption{Answer Capability Levels (ACLs) used to evaluate competency-question answerability.}
\label{tab:acl}
\begin{tabular}{p{0.07\linewidth} p{0.26\linewidth} p{0.59\linewidth}}
\toprule
\textbf{ACL} & \textbf{Capability} & \textbf{Interpretation} \\
\midrule
3 & Direct resolution & The answer is explicitly represented and directly retrievable. \\
2 & Computable resolution & The graph contains sufficient structure to derive the answer through traversal, rules, or deterministic computation. \\
1 & Structural retrieval & The graph identifies a constrained set of source material from which the answer can be obtained through additional human or language-model processing. \\
0 & No support & The schema or populated graph lacks the information required to assist with the question. \\
\bottomrule
\end{tabular}
\end{table}

Competency-question evaluations are typically binary, but binary answerability understates information coverage in SKGs. An answer may be explicitly represented in the graph or implicitly supported when the graph preserves and retrieves the source information from which a downstream computational component can derive it. We therefore introduce four \emph{Answer Capability Levels} (ACLs), ranging from no support~(0) to direct resolution~(3), as defined in \Cref{tab:acl}. Because a single use case may be operationalized through multiple competency questions, following Ontology Development 101~\cite{b_onto_101}, we define its cumulative satisfiability as the mean ACL across its question set \(Q_U\):

\begin{equation}
S_{\mathrm{UC}}(U)
=
\frac{1}{\lvert Q_U \rvert}
\sum_{q_i \in Q_U}
\mathrm{ACL}(q_i).
\label{eq:use-case-satisfiability}
\end{equation}

Averaging retains the \(0\)--\(3\) ACL scale, allowing the cumulative use-case score to preserve the same interpretation while supporting fair comparison across SKGs. Family-level coverage is obtained by macro-averaging the six use-case scores within each family.

\subsubsection{Experimental Setup}

We constructed model-specific competency-question sets for four SKGs under the proposed use-case collection. CS-KG~\cite{p_cs_kg} represents concept-centric modeling, while ORKG~\cite{p_orkg} and NLP-AKG~\cite{p_nlp_akg} represent statement-centric modeling through manual and automated population, respectively. These models cover the dominant SKG representation units and construction strategies. The CS-KG\footnote{\url{https://zenodo.org/records/14167682}} and ORKG\footnote{\url{https://orkg.org/data}} questions were evaluated against their publicly available knowledge-graph dumps. Because no comparable NLP-AKG dump was available, we generated one by applying its implementation to the ISWC publication collection used for SUDO. The resulting competency questions were evaluated using the ACL procedure defined above.

\subsubsection{Results and Discussion}
As shown in \Cref{fig:uc_analysis}, SUDO achieves the highest and most balanced coverage, with most use cases reaching ACL~2 or ACL~3 and substantial support maintained across all five families. CS-KG reaches ACL~3 for several explicitly modeled concept relations but falls to ACL~0 across most document-structure, argument, and pragmatic-description requirements, producing narrow but direct coverage. ORKG and NLP-AKG broaden coverage to contribution- and statement-oriented use cases, particularly those concerning claims and research contributions. Both reach ACL~3 when these elements are explicitly modeled, but decline to ACL~0 or ACL~1 when answering requires explicit connections among statements, paper-local concept roles, detailed document structure, or cross-publication context. ORKG retains intermediate support through its manually instantiated contribution templates, whereas NLP-AKG provides direct support where a requirement matches one of its predefined statement types but preserves little relevant information outside them.

SUDO’s stronger ACL profile follows from how its structures complement one another. Publication metadata and document structure directly resolve their corresponding requirements; Argument and Descriptor propositions capture research narratives and pragmatic detail; and the distinction between Artifacts and canonical Concepts supports both contextualized concept use and cross-publication connections. These structures produce ACL~3 when the requested answer is explicitly represented, ACL~2 when it can be derived from graph structure, and ACL~1 when source-grounded propositions preserve the information required for subsequent interpretation. SUDO’s coverage therefore reflects not only a broader explicit vocabulary but also its ability to retain answer-bearing information when no dedicated graph relation exists. These results support RQ1 by showing that SUDO represents scholarly knowledge more comprehensively across the full publication than the evaluated concept- or statement-centric models.

\subsection{RQ2: Utility for Scholarly Downstream Tasks} \label{sec:rq2}

\begin{table*}[t]
\centering
\caption{QASPER performance comparison across retrieval modalities and knowledge-graph systems.}
\label{tab:modality-performance}

\begingroup
\setlength{\tabcolsep}{5pt}
\renewcommand{\arraystretch}{1.0}

\begin{tabular}{
    ll
    S[table-format=3.0]
    S[table-format=2.2]
    S[table-format=2.2]
    S[table-format=2.2]
}
\toprule
\textbf{Modality} &
\textbf{Model} &
{\textbf{Test Size}} &
{\textbf{Precision}} &
{\textbf{Recall}} &
{\textbf{F1}} \\
\midrule

\multirow{3}{*}{Text~\cite{p_nlp_akg}}
    & BM25
    & 200
    & 67.94
    & 73.86
    & 70.84 \\

    & Embedding retrieval
    & 200
    & 68.29
    & 73.83
    & 70.69 \\

    & GPT-4
    & 200
    & 67.09
    & 72.20
    & 69.29 \\

\midrule

\multirow{3}{*}{NLP-AKG (Graph)~\cite{p_nlp_akg}}
    & KAG
    & 200
    & 68.73
    & 73.87
    & 70.97 \\

    & MindMap
    & 200
    & 69.38
    & 74.91
    & 71.73 \\

    & NLP-AKG (Agent)
    & 200
    & 69.43
    & 75.50
    & 72.04 \\

\midrule

CS-KG
    & \multicolumn{1}{c}{--}
    & \multicolumn{1}{c}{--}
    & \multicolumn{1}{c}{--}
    & \multicolumn{1}{c}{--}
    & \multicolumn{1}{c}{--} \\

ORKG
    & \multicolumn{1}{c}{--}
    & \multicolumn{1}{c}{--}
    & \multicolumn{1}{c}{--}
    & \multicolumn{1}{c}{--}
    & \multicolumn{1}{c}{--} \\

\midrule

\multirow{2}{*}{SUDO-KG}
    & Embedding retrieval
    & 600
    & 75.47
    & 81.11
    & 78.16 \\

    & gpt-oss-20b
    & 600
    & {\bfseries 82.57}
    & {\bfseries 82.39}
    & {\bfseries 82.44} \\

\bottomrule
\end{tabular}
\endgroup
\end{table*}

RQ2 examines whether SUDO’s discourse-aware representation improves access to answer-bearing information in scientific question answering, specifically in LLM-based scientific workflows. We evaluate 600 question--answer pairs from the QASPER dataset~\cite{d_qasper}, spanning 413 scientific publications and grounded in their source content. Because QASPER provides document-level questions with varying complexity, it suits evaluating scientific evidence retrieval without conflating results with open-ended agent planning. A SUDO knowledge graph is constructed for these publications using the same neuro-symbolic pipeline applied to the ISWC collection. We evaluate the retrieved information using BERTScore~\cite{p_bert} precision, recall, and F1: precision reflects the exclusion of unrelated content, recall reflects the retention of answer-relevant information, and F1 balances both.

\subsubsection{Results and Discussion}
As reported in \Cref{tab:modality-performance}, SUDO’s embedding-based retrieval achieves an F1 score of 78.16, exceeding the optimized NLP-AKG agent by approximately six percentage points. Query-guided retrieval further increases the F1 score to 82.44, with BERTScore precision and recall of 82.57 and 82.39, respectively, improving by about 10 percentage points over NLP-AKG. These results suggest that SUDO’s publication-local concepts and propositions provide more focused access to answer-bearing information than coarser statement units. The additional gain from query-guided retrieval further indicates that language models can use SUDO’s typed structure to refine information access.

Directly comparable scores could not be established for CS-KG or ORKG under the same evaluation protocol. CS-KG yielded too few reliable question-to-concept alignments for aggregate evaluation, reflecting the limited contextual information available from entity-level retrieval for publication-grounded questions. The available ORKG graph covered only 34 of the 413 QASPER publications, making a corpus-level comparison unrepresentative. These constraints reflect different modeling priorities: CS-KG emphasizes scalable concept-level organization, whereas ORKG provides richer contribution-level representations through manual curation, which constrains comprehensive population across large corpora. SUDO combines these strengths through scalable concept organization and source-grounded propositions, demonstrating the value of publication-centric, discourse-aware modeling for large-scale scientific question answering.

\section{Future Work}
Future work will investigate SUDO as an intermediary representation between publication text and specialized SKGs. Rather than extracting information directly into a predefined domain- or task-specific schema, a construction pipeline can first represent the publication through SUDO and derive specialized representations from that shared layer. Because the derived knowledge remains connected to SUDO’s Artifacts, Propositions, source locations, and provenance, specialized SKGs can reorganize or extend it without forfeiting source grounding. Concept-centric models can align their entities with SUDO Concepts while inheriting the publication-local roles captured by associated Artifacts and Propositions. Statement-centric models can construct higher-level claims and contributions over SUDO’s propositional structure while retaining traceable connections to the expressions from which they are derived. Future work will implement and evaluate the ontology alignments, identity mappings, and transformation pipelines required to materialize these complementary views.

This layered architecture also enables downstream applications that depend on how knowledge is expressed and connected. Within publications, it can support evidence tracing, limitation analysis, and claim verification; across publications, it can enable contribution comparison, method-lineage analysis, and investigation of how concepts and claims develop through the literature. For LLM-assisted workflows, SUDO can provide structured, source-grounded contexts for scientific question answering, knowledge discovery, literature-based hypothesis generation, and scholarly analysis. We will investigate domain- and task-specific extensions for these applications and evaluate their construction and use across larger scholarly collections. The broader goal is to enable scholarly systems to operate not only over extracted information, but also over how that information is introduced, justified, related, and propagated.

\section{Conclusion}
In this paper, we introduced SUDO, a domain-agnostic, discourse-aware ontology for representing scholarly knowledge through the complementary information and propositional structures of conceptual semantics. SUDO connects scholarly concepts, source-grounded propositions, research arguments, descriptive context, document structure, and provenance, preserving both how knowledge is communicated within a publication and how recurring concepts connect that knowledge across the literature to build global scholarly discourse structure. Evaluation across comprehensive scholarly knowledge-modeling use cases showed broader representational coverage than the examined concept- and statement-centric SKGs, while the QASPER experiment demonstrated a measurable advantage for retrieving focused, answer-bearing information in an LLM-based scientific question-answering workflow. By combining scalable concept organization with semantically rich propositions and precise source grounding, SUDO provides a shared foundation for specialized SKG extensions, cross-publication analysis, and future web-scale scholarly AI workflows.






\section*{Declaration on Generative AI}
  
 During the preparation of this work, the author(s) used GPT-5.5 and Gemini in order to: Grammar and spelling checks. After using these tool(s)/service(s), the author(s) reviewed and edited the content as needed and take(s) full responsibility for the publication’s content. 

\newpage
\bibliography{ref}

@inproceedings{p_semschol,
  title={Construction of the literature graph in semantic scholar},
  author={Ammar, Waleed and Groeneveld, Dirk and Bhagavatula, Chandra and Beltagy, Iz and Crawford, Miles and Downey, Doug and Dunkelberger, Jason and Elgohary, Ahmed and Feldman, Sergey and Ha, Vu and others},
  booktitle={Proceedings of the 2018 Conference of the North American Chapter of the Association for Computational Linguistics: Human Language Technologies, Volume 3 (Industry Papers)},
  pages={84--91},
  year={2018}
}

@inproceedings{p_ai_kg,
  title={AI-KG: an automatically generated knowledge graph of artificial intelligence},
  author={Dess{\`\i}, Danilo and Osborne, Francesco and Reforgiato Recupero, Diego and Buscaldi, Davide and Motta, Enrico and Sack, Harald},
  booktitle={International semantic web conference},
  pages={127--143},
  year={2020},
  organization={Springer}
}

@article{p_cs_kg2,
  title={Cs-kg 2.0: A large-scale knowledge graph of computer science},
  author={Dess{\'\i}, Danilo and Osborne, Francesco and Buscaldi, Davide and Reforgiato Recupero, Diego and Motta, Enrico},
  journal={Scientific Data},
  volume={12},
  number={1},
  pages={964},
  year={2025},
  publisher={Nature Publishing Group UK London}
}

@inproceedings{p_cs_kg,
  title={Cs-kg: A large-scale knowledge graph of research entities and claims in computer science},
  author={Dess{\'\i}, Danilo and Osborne, Francesco and Reforgiato Recupero, Diego and Buscaldi, Davide and Motta, Enrico},
  booktitle={International Semantic Web Conference},
  pages={678--696},
  year={2022},
  organization={Springer}
}

@inproceedings{p_mag,
  title={The microsoft academic knowledge graph: A linked data source with 8 billion triples of scholarly data},
  author={F{\"a}rber, Michael},
  booktitle={International semantic web conference},
  pages={113--129},
  year={2019},
  organization={Springer}
}

@inproceedings{p_orkg,
  title={Open research knowledge graph: a system walkthrough},
  author={Jaradeh, Mohamad Yaser and Oelen, Allard and Prinz, Manuel and Stocker, Markus and Auer, S{\"o}ren},
  booktitle={International Conference on Theory and Practice of Digital Libraries},
  pages={348--351},
  year={2019},
  organization={Springer}
}

@article{p_nlp_akg,
  title={Nlp-akg: Few-shot construction of nlp academic knowledge graph based on llm},
  author={Lan, Jiayin and Li, Jiaqi and Wang, Baoxin and Liu, Ming and Wu, Dayong and Wang, Shijin and Qin, Bing},
  journal={arXiv preprint arXiv:2502.14192},
  year={2025}
}

@article{p_openalex,
  title={OpenAlex: A fully-open index of scholarly works, authors, venues, institutions, and concepts},
  author={Priem, Jason and Piwowar, Heather and Orr, Richard},
  journal={arXiv preprint arXiv:2205.01833},
  year={2022}
}

@inproceedings{p_arnetminer,
  title={Arnetminer: extraction and mining of academic social networks},
  author={Tang, Jie and Zhang, Jing and Yao, Limin and Li, Juanzi and Zhang, Li and Su, Zhong},
  booktitle={Proceedings of the 14th ACM SIGKDD international conference on Knowledge discovery and data mining},
  pages={990--998},
  year={2008}
}

@article{p_mag2,
  title={Microsoft academic graph: When experts are not enough},
  author={Wang, Kuansan and Shen, Zhihong and Huang, Chiyuan and Wu, Chieh-Han and Dong, Yuxiao and Kanakia, Anshul},
  journal={Quantitative Science Studies},
  volume={1},
  number={1},
  pages={396--413},
  year={2020},
  publisher={MIT Press One Rogers Street, Cambridge, MA 02142-1209, USA journals-info~…}
}

@inproceedings{p_aminer,
  title={Name disambiguation in AMiner: Clustering, maintenance, and human in the loop.},
  author={Zhang, Yutao and Zhang, Fanjin and Yao, Peiran and Tang, Jie},
  booktitle={Proceedings of the 24th ACM SIGKDD international conference on knowledge discovery \& data mining},
  pages={1002--1011},
  year={2018}
}

@article{p_scholonto,
  title={ScholOnto: an ontology-based digital library server for research documents and discourse},
  author={Buckingham Shum, Simon and Motta, Enrico and Domingue, John},
  journal={International Journal on Digital Libraries},
  volume={3},
  number={3},
  pages={237--248},
  year={2000},
  publisher={Springer}
}

@article{p_semsur,
  title={SemSur: a core ontology for the semantic representation of research findings},
  author={Fathalla, Said and Vahdati, Sahar and Auer, S{\"o}ren and Lange, Christoph},
  journal={Procedia computer science},
  volume={137},
  pages={151--162},
  year={2018},
  publisher={Elsevier}
}

@article{d_spiqa,
  title={Spiqa: A dataset for multimodal question answering on scientific papers},
  author={Pramanick, Shraman and Chellappa, Rama and Venugopalan, Subhashini},
  journal={Advances in Neural Information Processing Systems},
  volume={37},
  pages={118807--118833},
  year={2024}
}

@article{d_sciqag,
  title={SciQAG: A framework for auto-generated science question answering dataset with fine-grained evaluation},
  author={Wan, Yuwei and Liu, Yixuan and Ajith, Aswathy and Grazian, Clara and Hoex, Bram and Zhang, Wenjie and Kit, Chunyu and Xie, Tong and Foster, Ian},
  journal={arXiv preprint arXiv:2405.09939},
  year={2024}
}

@inproceedings{d_qasper,
  title={A dataset of information-seeking questions and answers anchored in research papers},
  author={Dasigi, Pradeep and Lo, Kyle and Beltagy, Iz and Cohan, Arman and Smith, Noah A and Gardner, Matt},
  booktitle={Proceedings of the 2021 Conference of the North American Chapter of the Association for Computational Linguistics: Human Language Technologies},
  pages={4599--4610},
  year={2021}
}

@book{p_commitment_act,
  author    = {Robert B. Brandom},
  title     = {Making It Explicit: Reasoning, Representing, and Discursive Commitment},
  edition   = {4th},
  publisher = {Harvard University Press},
  year      = {1994},
  address   = {Cambridge, Mass.}
}

@book{p_speach_act,
  title     = {Pragmatics of Speech Actions},
  editor    = {Sbisà, Marina and Turner, Ken},
  year      = {2013},
  publisher = {De Gruyter Mouton},
  address   = {Berlin, Boston},
  doi       = {10.1515/9783110214383},
  url       = {https://doi.org/10.1515/9783110214383},
  urldate   = {2026-08-06},
  isbn      = {978-3-11-021438-3}
}

@techreport{p_rst,
  title={Rhetorical structure theory: A theory of text organization},
  author={Mann, William C and Thompson, Sandra A},
  year={1987},
  institution={University of Southern California, Information Sciences Institute Los Angeles}
}

@article{p_toulmin,
  title={Using Toulmin's model of argumentation},
  author={Karbach, Joan},
  journal={Journal of Teaching Writing},
  volume={6},
  number={1},
  pages={81--92},
  year={1987}
}

@article{p_kggen,
  title={Kggen: Extracting knowledge graphs from plain text with language models},
  author={Mo, Belinda and Yu, Kyssen and Kazdan, Joshua and Mpala, Proud and Yu, Lisa and Kanatsoulis, Charilaos and Koyejo, Sanmi},
  journal={Advances in Neural Information Processing Systems},
  volume={38},
  pages={30092--30115},
  year={2026}
}

@article{p_sci_disc_review,
  title={Literature based discovery: models, methods, and trends},
  author={Henry, Sam and McInnes, Bridget T},
  journal={Journal of biomedical informatics},
  volume={74},
  pages={20--32},
  year={2017},
  publisher={Elsevier}
}

@book{p_concept_sem,
  editor    = {Claudia Maienborn and Klaus von Heusinger and Paul Portner},
  title     = {Semantics - Theories},
  series    = {Mouton Reader},
  publisher = {De Gruyter Mouton},
  address   = {Berlin, Boston},
  year      = {2019},
  isbn      = {978-3110587036},
  doi       = {10.1515/9783110589245}
}

@inproceedings{p_deligent,
  title={Arguing on Issues with Mathematical Knowledge Items in a Semantic Wiki.},
  author={Lange, Christoph and Hastrup, Tuukka and Corlosquet, St{\'e}phane},
  booktitle={LWA},
  pages={41--48},
  year={2008}
}

@inproceedings{p_litsearch,
  title={Litsearch: A retrieval benchmark for scientific literature search},
  author={Ajith, Anirudh and Xia, Mengzhou and Chevalier, Alexis and Goyal, Tanya and Chen, Danqi and Gao, Tianyu},
  booktitle={Proceedings of the 2024 Conference on Empirical Methods in Natural Language Processing},
  pages={15068--15083},
  year={2024}
}

@article{p_alien_space,
  title={The Alien Space of Science: Sampling Coherent but Cognitively Unavailable Research Directions},
  author={Artiles, Alejandro H and Weiss, Martin and Brinkmann, Levin and Rahwan, Iyad and Sch{\"o}lkopf, Bernhard and Pal, Christopher and Larochelle, Hugo and Goyal, Anirudh and Rahaman, Nasim},
  journal={arXiv preprint arXiv:2603.01092},
  year={2026}
}

@article{p_ai_scien,
  title={The ai scientist: Towards fully automated open-ended scientific discovery},
  author={Lu, Chris and Lu, Cong and Lange, Robert Tjarko and Foerster, Jakob and Clune, Jeff and Ha, David},
  journal={arXiv preprint arXiv:2408.06292},
  year={2024}
}

@article{p_alpha_evolve,
  title={Alphaevolve: A coding agent for scientific and algorithmic discovery},
  author={Novikov, Alexander and V{\~u}, Ng{\^a}n and Eisenberger, Marvin and Dupont, Emilien and Huang, Po-Sen and Wagner, Adam Zsolt and Shirobokov, Sergey and Kozlovskii, Borislav and Ruiz, Francisco JR and Mehrabian, Abbas and others},
  journal={arXiv preprint arXiv:2506.13131},
  year={2025}
}

@article{p_ai_scien2,
  title={The ai scientist-v2: Workshop-level automated scientific discovery via agentic tree search},
  author={Yamada, Yutaro and Lange, Robert Tjarko and Lu, Cong and Hu, Shengran and Lu, Chris and Foerster, Jakob and Clune, Jeff and Ha, David},
  journal={arXiv preprint arXiv:2504.08066},
  year={2025}
}

@book{b_dst,
  title={How to do things with words},
  author={Austin, John Langshaw},
  volume={5},
  year={1975},
  publisher={Harvard university press}
}

@book{p_illocutionary,
  title={Studies in the Way of Words},
  author={Grice, Paul},
  year={1991},
  publisher={Harvard University Press}
}

@incollection{p_dst,
  title={Segmented discourse representation theory: Dynamic semantics with discourse structure},
  author={Lascarides, Alex and Asher, Nicholas},
  booktitle={Computing meaning},
  pages={87--124},
  year={2007},
  publisher={Springer}
}

@book{p_illoc_logic,
  title={Foundations of illocutionary logic},
  author={Searle, John R and Vanderveken, Daniel},
  year={1985},
  publisher={CUP Archive}
}

@article{p_attention,
  title={Attention is all you need},
  author={Vaswani, Ashish and Shazeer, Noam and Parmar, Niki and Uszkoreit, Jakob and Jones, Llion and Gomez, Aidan N and Kaiser, {\L}ukasz and Polosukhin, Illia},
  journal={Advances in neural information processing systems},
  volume={30},
  year={2017}
}

@inproceedings{p_bert,
  title={Bert: Pre-training of deep bidirectional transformers for language understanding},
  author={Devlin, Jacob and Chang, Ming-Wei and Lee, Kenton and Toutanova, Kristina},
  booktitle={Proceedings of the 2019 conference of the North American chapter of the association for computational linguistics: human language technologies, volume 1 (long and short papers)},
  pages={4171--4186},
  year={2019}
}

@article{p_found_lang,
  title={Pr{\'e}cis of foundations of language: Brain, meaning, grammar, evolution},
  author={Jackendoff, Ray},
  journal={Behavioral and brain sciences},
  volume={26},
  number={6},
  pages={651--665},
  year={2003},
  publisher={Cambridge University Press}
}

@phdthesis{p_arg_zone_thesis,
  title={Argumentative zoning: Information extraction from scientific text},
  author={Teufel, Simone and others},
  year={1999},
  school={University of Edinburgh Edinburgh}
}

@inproceedings{p_arg_zoning,
  title = {Segmentation of {{Argumentative Texts}} by {{Key Statements}} for {{Argument Mining}} from the {{Web}}},
  booktitle = {Proceedings of the 12th {{Argument}} Mining {{Workshop}}},
  author = {Zelch, Ines and Hagen, Matthias and Stein, Benno and Kiesel, Johannes},
  editor = {Chistova, Elena and Cimiano, Philipp and Haddadan, Shohreh and Lapesa, Gabriella and {Ruiz-Dolz}, Ramon},
  year = 2025,
  month = jul,
  pages = {228--242},
  publisher = {Association for Computational Linguistics},
  address = {Vienna, Austria},
  doi = {10.18653/v1/2025.argmining-1.22},
  urldate = {2026-01-31},
  isbn = {979-8-89176-258-9}
}

@inproceedings{p_toward_arg_zone,
  title = {Towards {{Domain-Independent Argumentative Zoning}}: {{Evidence}} from {{Chemistry}} and {{Computational Linguistics}}},
  shorttitle = {Towards {{Domain-Independent Argumentative Zoning}}},
  booktitle = {Proceedings of the 2009 {{Conference}} on {{Empirical Methods}} in {{Natural Language Processing}}},
  author = {Teufel, Simone and Siddharthan, Advaith and Batchelor, Colin},
  editor = {Koehn, Philipp and Mihalcea, Rada},
  year = 2009,
  month = aug,
  pages = {1493--1502},
  publisher = {Association for Computational Linguistics},
  address = {Singapore},
  urldate = {2026-01-31}
}

@article{p_owl_onto,
  title={OWL 2 web ontology language document overview},
  author={World Wide Web Consortium and others},
  year={2012},
  publisher={Word Wide Web Consortium}
}

@book{b_onto_101,
  title={Ontology engineering},
  author={Kendall, Elisa F and McGuinness, Deborah L},
  year={2019},
  publisher={Morgan \& Claypool Publishers}
}

@article{p_oops, 
	title={{OOPS! (OntOlogy Pitfall Scanner!): An On-line Tool for Ontology Evaluation}},
	author={Poveda-Villal{\'o}n, Mar{\'i}a and G{\'o}mez-P{\'e}rez, Asunci{\'o}n and Su{\'a}rez-Figueroa, Mari Carmen},
	journal={International Journal on Semantic Web and Information Systems (IJSWIS)},
	volume={10}, 
	number={2}, 
	pages={7--34}, 
	year={2014}, 
	publisher={IGI Global} 
	}

@inproceedings{p_widoco,
  title={WIDOCO: a wizard for documenting ontologies},
  author={Garijo, Daniel},
  booktitle={International Semantic Web Conference},
  pages={94--102},
  year={2017},
  organization={Springer, Cham},
  doi = {10.1007/978-3-319-68204-4_9},
  funding = {USNSF ICER-1541029, NIH 1R01GM117097-01},
  url={http://dgarijo.com/papers/widoco-iswc2017.pdf}
}

@article{p_disc_skg,
  title={Steps towards an infrastructure for scholarly synthesis},
  author={Chan, Joel and Akamatsu, Matthew and Vargas, David and Kawerau, Lukas and Gartner, Michael},
  journal={arXiv preprint arXiv:2407.20666},
  year={2024}
}

@misc{o_amo,
  author       = {Peroni, Silvio and Vitali, Fabio},
  title        = {The {Argument Model Ontology} ({AMO})},
  howpublished = {\url{https://sparontologies.github.io/amo/current/amo.html}},
  year         = {2011},
  note         = {Accessed: 2026-09-16}
}

\newpage
\appendix

\section{Appendices}



\subsection{Scholarly Publication Knowledge-Representation Use Cases} \label{app:skg_use_cases}

\begingroup
\setlength{\LTleft}{0pt}
\setlength{\LTright}{0pt}
\setlength{\tabcolsep}{4pt}
\renewcommand{\arraystretch}{1.12}
\footnotesize

\begin{longtable}{
    @{}
    P{0.07\linewidth}
    P{0.20\linewidth}
    P{0.67\linewidth}
    @{}
}
\caption{Scholarly Publication Knowledge-Representation Use Cases}
\label{tab:publication-use-cases}\\

\toprule
\textbf{ID} &
\textbf{Title} &
\textbf{Description} \\
\midrule
\endfirsthead

\multicolumn{3}{@{}l}{
    \footnotesize\textit{Table \thetable{} continued from the previous page}
}\\
\toprule
\textbf{ID} &
\textbf{Title} &
\textbf{Description} \\
\midrule
\endhead

\midrule
\multicolumn{3}{r@{}}{
    \footnotesize\textit{Continued on the next page}
}\\
\endfoot

\bottomrule
\endlastfoot


\multicolumn{3}{@{}l@{}}{%
    \textbf{Family D: Publication Metadata and Document Structure}%
}\\[0.35em]

UC-D1 &
Identity &
Which bibliographic attributes, such as title, DOI, publication date,
venue, publication type, and license, identify a given publication? \\[0.35em]

UC-D2 &
Agent Role &
Which agents contributed to a publication, and what explicit structural
or scholarly roles, such as lead author, corresponding author, editor,
reviewer, or data curator, did they perform? \\[0.35em]

UC-D3 &
Associations &
What directional relationships connect a publication to external
scholarly artifacts, such as preprints, datasets, software,
supplementary materials, errata, or citations, and how are these
relationships typed? \\[0.35em]

UC-D4 &
Structural Composition &
What hierarchical and sequential document components, such as sections,
subsections, paragraphs, sentences, and figures, compose a publication,
and what is their absolute linear ordering? \\[0.35em]

UC-D5 &
Rhetorical Function &
What structural or rhetorical block type, such as background,
methodology, results, or limitations, characterizes each document
component of a publication? \\[0.35em]

UC-D6 &
Source Grounding &
Which exact document component and character- or byte-offset span
grounds a given extracted entity, claim, or relation within a
publication? \\

\midrule


\multicolumn{3}{@{}l@{}}{%
    \textbf{Family A: Paper Argument and Discourse}%
}\\[0.35em]

UC-A1 &
Objectives &
What specific research problems, questions, or objectives are explicitly
declared as the target of investigation in a publication? \\[0.35em]

UC-A2 &
Claims &
What distinct claims, hypotheses, findings, or core contributions are
advanced by a publication, and what are their fine-grained
propositional types? \\[0.35em]

UC-A3 &
Evidence Grounding &
What specific empirical evidence, observations, results, figures,
tables, or secondary citations ground or substantiate a given claim? \\[0.35em]

UC-A4 &
Discourse Relations &
What rhetorical or argumentative relations, such as supporting,
challenging, qualifying, explaining, motivating, or elaborating, exist
between distinct propositions within a publication? \\[0.35em]

UC-A5 &
Constraints &
What epistemic modal markers, assumptions, qualifications, or boundary
limitations constrain the validity of a given claim? \\[0.35em]

UC-A6 &
Argument Paths &
What directed path of discourse relations structurally connects the
initial research problem of a publication through its methods and
evidence to its final conclusions? \\

\midrule


\multicolumn{3}{@{}l@{}}{%
    \textbf{Family C: Concepts and Paper-local Use}%
}\\[0.35em]

UC-C1 &
Mentions &
What distinct textual expressions, surface forms, abbreviations, or
coreferential expressions mention a concept within a publication? \\[0.35em]

UC-C2 &
Global Typing &
To what canonical domain entity or global knowledge-base URI does a
concept mention resolve, and what is its fixed ontological type, such
as model, dataset, task, metric, theory, material, or algorithm? \\[0.35em]

UC-C3 &
Local Operational Role &
What dynamic, paper-local role, such as proposed method, baseline
approach, evaluation dataset, experimental variable, or output, does a
concept fulfill within the specific methodology of a publication? \\[0.35em]

UC-C4 &
Propositional Argument Roles &
In which specific semantic triples or n-ary propositions within a
publication does a concept participate, and what structural position
does it occupy? \\[0.35em]

UC-C5 &
Local Concept Alignment &
How is a concept contextually paired with or related to other concepts,
such as mapping a specific evaluation metric to a specific dataset,
within the experimental scope of a publication? \\[0.35em]

UC-C6 &
Pragmatic Disposition &
What is the authors' explicit disposition toward a concept, such as
introducing, adopting, reusing, extending, modifying, comparing,
evaluating, criticizing, or replacing it? \\

\midrule


\multicolumn{3}{@{}l@{}}{%
    \textbf{Family X: Cross-publication Concept Dynamics}%
}\\[0.35em]

UC-X1 &
Cross-paper Co-occurrence &
Which distinct publications reference the same canonical domain
concept, irrespective of local naming variations? \\[0.35em]

UC-X2 &
Role Shift Matrix &
How does the localized operational role of a single concept vary,
transition, or contrast across a collection of different publications? \\[0.35em]

UC-X3 &
Definitional Variance &
How do the formal definitions or textual characterizations of a
specific concept differ across distinct publications? \\[0.35em]

UC-X4 &
Dialectical Claims &
What epistemic alignments or conflicts, such as agreement,
contradiction, refinement, or qualification, exist between claims made
about the same concept across different papers? \\[0.35em]

UC-X5 &
Inter-paper Influence &
How does the treatment of a concept in one publication directly
influence, inherit from, reproduce, extend, or critique its treatment
in an earlier publication? \\[0.35em]

UC-X6 &
Evolutionary Trajectory &
What is the chronological trajectory of a concept's roles, usage
frequencies, and domain shifts across a temporal or community-based
graph projection? \\

\midrule


\multicolumn{3}{@{}l@{}}{%
    \textbf{Family P: Paper-local Pragmatic Descriptions}%
}\\[0.35em]

UC-P1 &
Intensional Definition &
How does the text of a publication intensionally define or describe
the essential nature of a concept? \\[0.35em]

UC-P2 &
Functional Intent &
What explicit objective or goal-driven motivation does the text of a
publication assign to the selection or deployment of a concept? \\[0.35em]

UC-P3 &
Operationalization Parameters &
How is a concept configured, parameterized, operationalized, or
structured into the systemic workflow described in a publication? \\[0.35em]

UC-P4 &
Attributed Affordances &
What explicit capabilities, advantages, performance benefits, or
unique affordances does a publication attribute to a concept? \\[0.35em]

UC-P5 &
Boundary Conditions &
Under what explicit environmental constraints, domain dependencies,
assumptions, or operational limits does a publication declare a concept
to be functional? \\[0.35em]

UC-P6 &
Pragmatic Contrast &
How does a publication contrast the utility, trade-offs, or suitability
of a concept against alternative baselines or rival concepts? \\

\end{longtable}

\endgroup

\subsection{Online Resources} 
\label{app:online-resources}

\begin{description}
    \item[SUDO ontology:]
    \url{https://w3id.org/twc/sudo/ }

    \item[SUDO documentation:]
    \url{https://tetherless-world.github.io/sudo-ontology/}
    
    \item[ISWC publication corpus:]
    \url{https://zenodo.org/records/21812666}

    \item[ISWC SUDO knowledge graph:]
    \url{https://zenodo.org/records/21813306}

    \item[QASPER SUDO knowledge graph:]
    \url{https://zenodo.org/records/21815806}
\end{description}

\subsection{Concept-Structure Classes and Relations}
\label{app:concept_structure}

SUDO defines five domain-agnostic subclasses of \textbf{Artifact} to capture the broad semantic categories through which scholarly nomenclature is organized. Their definitions and representative examples are summarized in \Cref{tab:artifact_classes}.

\begin{table}[!h]
\centering
\caption{Artifact classes used in the SUDO concept structure.}
\label{tab:artifact_classes}
\begin{tabularx}{\linewidth}{@{}l X X@{}}
\toprule
\textbf{Class} & \textbf{Definition} & \textbf{Examples} \\
\midrule
\textbf{Task} &
A problem to solve, application to support, or system to construct. &
Information extraction, machine reading, image segmentation. \\

\textbf{Method} &
A method, model, system, tool, framework, or system component. &
Language model, CoreNLP, POS parser, kernel method. \\

\textbf{Metric} &
A measure used to express the quality, efficiency, or behavior of a method or system. &
F1, BLEU, precision, recall, robustness, time complexity. \\

\textbf{Material} &
Data, datasets, corpora, resources, or knowledge bases used in research. &
Image data, bilingual dictionary, CoNLL corpus, paraphrased questions. \\

\textbf{Term} &
A significant scholarly notion that does not fall under the preceding categories. &
Hallucination, faithfulness, Rhetorical Structure Theory, named entity. \\
\bottomrule
\end{tabularx}
\end{table}

Relations among Artifacts capture domain-independent functional, taxonomic, and compositional associations. SUDO also reuses SKOS properties to align canonical concepts with related instances and lexical variants, as summarized in \Cref{tab:concept_relations}.







\begin{table}[!h]
\centering
\caption{Relations used in the SUDO concept structure.}
\label{tab:concept_relations}
\begin{tabularx}{\linewidth}{@{}l X X@{}}
\toprule
\textbf{Relation} & \textbf{Interpretation} & \textbf{Example} \\
\midrule

\textbf{used-for} &
Indicates that one Artifact is used to perform, support, or construct another. &
\textbf{TISPER} \textit{is used for} \textbf{text-processing applications}. \\

\textbf{feature-of} &
Indicates that one Artifact is a feature, property, or characteristic of another. &
\textbf{Genre-specific regularities} \textit{are features of} \textbf{discourse structure}. \\

\textbf{hyponym-of} &
Expresses a taxonomic relation in which one Artifact is a more specific type of another. &
\textbf{TUIT} \textit{is a type of} \textbf{software library}. \\

\textbf{part-of} &
Expresses a compositional relation in which one Artifact forms a component of another. &
A \textbf{natural-language understanding module} \textit{is part of} a \textbf{dialogue system}. \\

\textbf{closeMatch} &
Connects publication-specific Artifacts to canonical Concepts with sufficiently similar meanings, without asserting strict identity. &
The \textbf{Transformer Artifact} in each publication \textit{closely matches} the canonical \textbf{Transformer Concept}. \\

\textbf{altLabel} &
Associates a canonical Concept with an alternative lexical form, abbreviation, or spelling variant. &
The \textbf{BERT Concept} \textit{has the alternative label} \textbf{Bidirectional Encoder Representations from Transformers}. \\

\bottomrule
\end{tabularx}
\end{table}

\subsection{Ontology Reuse and Vocabulary}
\label{app:ontology_vocabulary}

SUDO reuses established vocabularies for publication metadata, document
structure, knowledge organization, and argumentation. 
\Cref{tab:reused_ontologies} lists these ontologies, while
\Cref{tab:sudo_vocabulary} summarizes the classes, datatypes, and properties
used in each representational layer.

\begin{longtable}{
    @{}
    >{\raggedright\arraybackslash}p{0.12\textwidth}
    >{\raggedright\arraybackslash}p{0.39\textwidth}
    >{\raggedright\arraybackslash}p{0.41\textwidth}
    @{}
}
\caption{Ontologies and vocabularies reused or referenced by SUDO.}
\label{tab:reused_ontologies}\\
\toprule
\textbf{Prefix} & \textbf{Ontology or vocabulary} & \textbf{Specification} \\
\midrule
\endfirsthead

\multicolumn{3}{l}{\small\textit{Table~\thetable\ continued}}\\
\toprule
\textbf{Prefix} & \textbf{Ontology or vocabulary} & \textbf{Specification} \\
\midrule
\endhead

\midrule
\multicolumn{3}{r}{\small\textit{Continued on the next page}}\\
\endfoot

\bottomrule
\endlastfoot

\ontterm{foaf} &
Friend of a Friend &
\url{https://xmlns.com/foaf/spec/} \\

\ontterm{bibo} &
The Bibliographic Ontology &
\url{https://dcmi.github.io/bibo/} \\

\ontterm{xsd} &
XML Schema Datatypes &
\url{https://www.w3.org/TR/swbp-xsch-datatypes/} \\

\ontterm{dcterms} &
Dublin Core Metadata Terms &
\url{https://www.dublincore.org/specifications/dublin-core/dcmi-terms/} \\

\ontterm{cito} &
The Citation Typing Ontology &
\url{https://sparontologies.github.io/cito/current/cito.html} \\

\ontterm{fabio} &
The FRBR-aligned Bibliographic Ontology &
\url{https://sparontologies.github.io/fabio/current/fabio.html} \\

\ontterm{doco} &
The Document Components Ontology &
\url{https://sparontologies.github.io/doco/current/doco.html} \\

\ontterm{po} &
The Pattern Ontology &
\url{https://sparontologies.github.io/po/current/po.html} \\

\ontterm{pav} &
Provenance, Authoring and Versioning Ontology &
\url{https://pav-ontology.github.io/pav/} \\

\ontterm{prov} &
The Provenance Ontology &
\url{https://www.w3.org/TR/prov-o/} \\

\ontterm{rdfs} &
RDF Schema &r
\url{https://www.w3.org/TR/rdf-schema/} \\

\ontterm{schema} &
Schema.org &
\url{https://schema.org/} \\

\ontterm{skos} &
Simple Knowledge Organization System &
\url{https://www.w3.org/TR/skos-reference/} \\

\ontterm{amo} &
The Argument Model Ontology &
\url{https://sparontologies.github.io/amo/current/amo.html} \\

\ontterm{swan} &
Semantic Web Applications in Neuromedicine Ontology &
\url{https://www.w3.org/TR/hcls-swan/} \\

\ontterm{semsur} &
Semantic Survey Ontology &
\url{https://saidfathalla.github.io/SemSur/doc/} \\

\ontterm{sudo} &
Scholarly Upper Discourse Ontology &
\url{https://w3id.org/twc/sudo/ontology} \\

\end{longtable}

\begin{longtable}{
    @{}
    >{\raggedright\arraybackslash}p{0.12\textwidth}
    >{\raggedright\arraybackslash}p{0.40\textwidth}
    >{\raggedright\arraybackslash}p{0.40\textwidth}
    @{}
}
\caption{Classes, datatypes, and properties used in the SUDO ontology,
grouped by representational layer.}
\label{tab:sudo_vocabulary}\\
\toprule
\textbf{Source} & \textbf{Classes and datatypes} & \textbf{Properties} \\
\midrule
\endfirsthead

\multicolumn{3}{l}{\small\textit{Table~\thetable\ continued}}\\
\toprule
\textbf{Source} & \textbf{Classes and datatypes} & \textbf{Properties} \\
\midrule
\endhead

\midrule
\multicolumn{3}{r}{\small\textit{Continued on the next page}}\\
\endfoot

\bottomrule
\endlastfoot

\multicolumn{3}{@{}l}{\textbf{Publication metadata}}\\
\addlinespace[2pt]

\ontterm{foaf} &
\ontterm{Person} &
--- \\

\ontterm{bibo} &
\ontterm{Proceedings} &
\ontterm{doi} \\

\ontterm{xsd} &
\ontterm{integer}, \ontterm{anyURI}, \ontterm{string} &
--- \\

\ontterm{dcterms} &
--- &
\ontterm{creator}, \ontterm{isPartOf}, \ontterm{subject},
\ontterm{title}, \ontterm{issued} \\

\ontterm{cito} &
--- &
\ontterm{cites} \\

\ontterm{rdfs} &
\ontterm{Literal} &
\ontterm{seeAlso} \\

\ontterm{sudo} &
--- &
\ontterm{citationCount} \\

\addlinespace
\multicolumn{3}{@{}l}{\textbf{Document structure}}\\
\addlinespace[2pt]

\ontterm{fabio} &
\ontterm{ResearchPaper} &
--- \\

\ontterm{doco} &
\ontterm{DiscourseElement}, \ontterm{Section},
\ontterm{SectionTitle}, \ontterm{Paragraph}, \ontterm{Sentence} &
--- \\

\ontterm{po} &
--- &
\ontterm{isContainedByAsHeader}, \ontterm{contains} \\

\ontterm{schema} &
--- &
\ontterm{position} \\

\addlinespace
\multicolumn{3}{@{}l}{\textbf{Information structure}}\\
\addlinespace[2pt]

\ontterm{skos} &
--- &
\ontterm{closeMatch}, \ontterm{altLabel} \\

\ontterm{sudo} &
\ontterm{Concept}, \ontterm{Artifact}, \ontterm{Material},
\ontterm{Method}, \ontterm{Metric}, \ontterm{Task}, \ontterm{Term} &
\ontterm{usedFor}, \ontterm{featureOf},
\ontterm{hyponymOf}, \ontterm{partOf} \\

\addlinespace
\multicolumn{3}{@{}l}{\textbf{Propositional structure: argumentation}}\\
\addlinespace[2pt]

\ontterm{amo} &
\ontterm{Argument}, \ontterm{Backing}, \ontterm{Claim},
\ontterm{Evidence}, \ontterm{Qualifier}, \ontterm{Rebuttal},
\ontterm{Warrant} &
\ontterm{backs}, \ontterm{forces}, \ontterm{isValidUnless},
\ontterm{leadsTo}, \ontterm{proves}, \ontterm{supports} \\

\ontterm{swan} &
\ontterm{Idea}, \ontterm{Issue} &
\ontterm{responseTo} \\

\ontterm{semsur} &
\ontterm{Approach} &
--- \\

\ontterm{sudo} &
--- &
\ontterm{motivates}, \ontterm{evidence}, \ontterm{contextualizes},
\ontterm{enables}, \ontterm{justify}, \ontterm{concede} \\

\addlinespace
\multicolumn{3}{@{}l}{\textbf{Propositional structure: description}}\\
\addlinespace[2pt]

\ontterm{sudo} &
\ontterm{Proposition}, \ontterm{Descriptor},
\ontterm{Elaboration}, \ontterm{Circumstance},
\ontterm{Solutionhood}, \ontterm{Cause}, \ontterm{Setting},
\ontterm{Interpretation}, \ontterm{Restatement},
\ontterm{Evaluation}, \ontterm{Purpose}, \ontterm{Summary},
\ontterm{Result}, \ontterm{Condition} &

\ontterm{contrast}, \ontterm{contextualizes}, \ontterm{enables},  \ontterm{summarizes}, \ontterm{result},
\ontterm{condition}, \ontterm{restates}, \ontterm{evaluates},
\ontterm{purposes}, \ontterm{causes}, \ontterm{setting},
\ontterm{interprets}, \ontterm{elaborates}, \ontterm{solves} \\

\addlinespace
\multicolumn{3}{@{}l}{\textbf{Provenance}}\\
\addlinespace[2pt]

\ontterm{prov} &
\ontterm{Agent} &
\ontterm{hadPrimarySource}, \ontterm{wasDerivedFrom} \\

\ontterm{pav} &
--- &
\ontterm{createdBy} \\

\end{longtable}

\newpage

\end{document}